\documentclass[conference]{IEEEtran}
\IEEEoverridecommandlockouts
\usepackage{cite}
\usepackage{amsmath,amssymb,amsfonts}
\usepackage{algorithmic}
\usepackage{graphicx}
\usepackage{textcomp}
\usepackage{xcolor}
\usepackage{eso-pic}
\usepackage{subcaption}
\usepackage{amssymb}
\usepackage{amsmath}
\usepackage{booktabs}
\usepackage{multirow}
\usepackage{graphicx} 
\usepackage{xcolor}
\usepackage{soul}
\usepackage{marginnote}

\definecolor{lightyellow}{RGB}{255, 255, 204}
\sethlcolor{lightyellow}

\def\BibTeX{{\rm B\kern-.05em{\sc i\kern-.025em b}\kern-.08em
    T\kern-.1667em\lower.7ex\hbox{E}\kern-.125emX}}
\begin{document}

\title{Pedestrian Crossing Intent Prediction via Psychological Features and Transformer Fusion}


\author{
\IEEEauthorblockN{Sima Ashayer, Hoang H. Nguyen, Yu Liang, Mina Sartipi}
\IEEEauthorblockA{The University of Tennessee at Chattanooga, Chattanooga, TN, USA\\
npk875@mocs.utc.edu, huuhoang-nguyen@utc.edu, yu-liang@utc.edu, mina-sartipi@utc.edu}
}

\maketitle

\begin{abstract}
Pedestrian intention prediction needs to be accurate for autonomous vehicles to navigate safely in urban environments. We present a lightweight, socially informed architecture for pedestrian intention prediction. It fuses four behavioral streams (attention, position, situation, and interaction) using highway encoders, a compact 4-token Transformer, and global self-attention pooling. To quantify uncertainty, we incorporate two complementary heads: a variational bottleneck whose KL divergence captures epistemic uncertainty, and a Mahalanobis distance detector that identifies distributional shift. Together, these components yield calibrated probabilities and actionable risk scores without compromising efficiency. On the PSI 1.0 benchmark, our model outperforms recent vision–language models by achieving 0.9 F1, 0.94 AUC-ROC, and 0.78 MCC by using only structured, interpretable features. On the more diverse PSI 2.0 dataset, where, to the best of our knowledge, no prior results exist, we establish a strong initial baseline of 0.78 F1 and 0.79 AUC-ROC. Selective prediction based on Mahalanobis scores increases test accuracy by up to 0.4 percentage points at 80\% coverage. Qualitative attention heatmaps further show how the model shifts its cross-stream focus under ambiguity. The proposed approach is modality-agnostic, easy to integrate with vision–language pipelines, and suitable for risk-aware intent prediction on resource-constrained platforms.
\end{abstract}


\section{Introduction}
Understanding and anticipating pedestrian behavior is a central challenge for autonomous driving systems, particularly in today’s complex urban environments where safe and reliable navigation is essential. Accurately identifying a pedestrian’s intention to cross, before the first step into the roadway, is critical for timely AV decision-making and collision prevention\cite{Uziel_2025_WACV}. Importantly, crossing intention is distinct from crossing prediction: the former reflects a pedestrian’s internal decision to cross, while the latter forecasts future presence on the road\cite{Uziel_2025_WACV}. This distinction underscores the cognitive aspect of pedestrian behavior, which AVs must infer in real time. Misjudging intent can cause either unsafe actions or unnecessary hesitation, both increasing accident risk \cite{lian2025dual}.

Several publicly available datasets have captured this nuance, including PIE \cite{rasouli2019pie}, JAAD \cite{rasouli2017they}, and more recently, PSI \cite{chen2021psi}, which provides dynamic intent annotations and natural language explanations from human drivers, offering a richer view of human-in-the-loop reasoning.\cite{chen2021psi}. Predicting pedestrian behavior is both a technical and societal challenge, as human drivers intuitively interpret cues such as gaze, posture, and traffic context to judge intent \cite{lian2025dual}. While AVs must replicate this reasoning algorithmically under strict safety and efficiency constraints \cite{lian2025dual,SHARMA2025111205}. This task is challenging because traffic flow, social interactions, and internal states can cause sudden behavioral changes \cite{lian2025dual,SHARMA2025111205}. Early trajectory-based methods \cite{sadeghian2019sophie} struggle with long-term uncertainty, while vision-based or skeleton-based models remain sensitive to occlusion and overlook contextual or social cues \cite{lian2025dual,guo20223d}.

To address these limitations, the Pedestrian Situated Intent (PSI) perspective views intention as a dynamic state shaped by behavior and context, which aligns with the idea that driving is both technical and social \cite{chen2021psi,pelikan2021autonomous}. AVs must therefore perceive and interpret the world, inferring the intentions of human road users collaboratively \cite{rasouli2019pie}. Infrastructure-assisted sensing and communication, such as infrastructure-to-pedestrian (I2P) systems, can complement perception-based intent models by detecting pedestrians in real time \cite{khaleghian2025software}. To support trust and safety, it is useful to include psychological (why), human-centered (for whom), and causal (what led to this) perspectives when designing intent models \cite{atakishiyev2024explainable}. The PSI dataset includes intent annotations during real-world pedestrian–vehicle interactions, along with verbal explanations from human drivers \cite{chen2021psi}. This combination offers an opportunity to train models that detect crossing intent and align better with how humans interpret behavior in complex urban environments.

We propose a socially informed, multi-stream architecture for pedestrian crossing intent prediction on PSI~1.0 and PSI~2.0. Our approach fuses psychological, spatial, situational, and interaction cues through a Transformer that attends across modalities. Robustness is improved using MixUp augmentation, label smoothing, and KL/Mahalanobis anomaly detection. PSI’s structured annotations, including driver explanations, help guide feature design and narrow the gap between human judgment and machine perception in autonomous driving. While our architecture uses established components (highway encoders, Transformer, residual MLP), the novelty lies in how these modules support psychologically grounded behavioral reasoning. We introduce a semantically organized 4-token representation to capture distinct aspects of pedestrian–vehicle interaction. Cross-stream attention over this compact representation enables calibrated and risk-aware prediction. Unlike prior models relying on deep temporal models or graph-based reasoning, our design prioritizes simplicity and real-time applicability. This framework offers a principled and extensible foundation for future temporal or relational modeling.

\section{Related work}
Early approaches to pedestrian crossing intent prediction used multi-branch models that fused visual and kinematic cues. Each modality (pose, location, and vehicle speed) was processed in separate streams and combined through attention or sequential fusion blocks. Ham et al. \cite{ham2023cipf} proposed a recurrent multi-stream model integrating pedestrian and vehicle features.

Such architectures often treat streams independently, which can limit cross-modal reasoning. Transformer-based models address this by learning unified temporal and cross-modal relationships. Zhou et al. \cite{10247098} introduced a Transformer architecture that encodes pedestrian and environmental elements as tokens to capture their dynamics. while Sharma et al. \cite{SHARMA2025111205} introduced IntentFormer, a multimodal Transformer for RGB, semantic, and trajectory inputs. These approaches underscore the benefits of joint representation learning.

Uncertainty estimation models further address ambiguity. Zhang et al. \cite{zhang2023trep} applied evidential deep learning within a Transformer to quantify prediction confidence for deployment under noisy or subjective labels. An emerging direction leverages foundation models. ClipCross~\cite{Uziel_2025_WACV} adapts CLIP embeddings for distinguishing crossing versus non-crossing, while Azarmi et al.~\cite{inproceedings} extend vision-language reasoning to video and prompt-based intent recognition. Zambare et al.~\cite{zambare2025seeing} further explored zero-shot generalization using Gemini~2.5.

Collectively, the field has evolved from modular pipelines to unified, socially informed models. We build on this trajectory with a multi-stream Transformer that encodes psychological attention, positional, situational, and interaction cues, enabled by the PSI dataset to enhance prediction and interpretability.

\begin{figure*}[ht]
    \centering
    \includegraphics[width=0.85\linewidth]{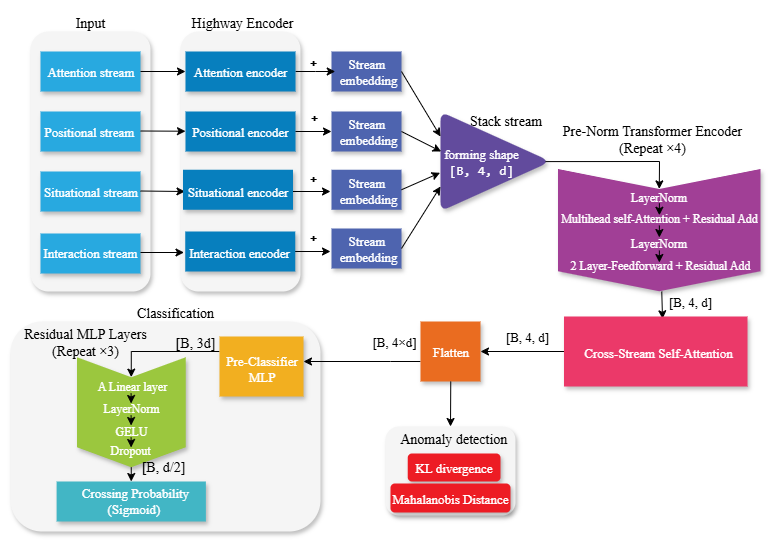}
    \caption{\textbf{Overview of socially informed, multi-stream architecture for pedestrian crossing intention prediction:} Four input streams—attention, positional, situational, and interaction features—are encoded by highway encoders and augmented with stream embeddings. A Transformer model with temporal and cross-stream dependencies and self-attention yields a unified representation. A residual MLP predicts crossing probability, while KL and Mahalanobis modules provide uncertainty and anomaly scores.}
    \label{fig:architecture}
\end{figure*}

\section{Datasets and Feature Engineering}
We used PSI~1.0/2.0 datasets \cite{chen2021psi,jing2022inaction}, which capture real-world interactions between vehicles and pedestrians in urban scenes. PSI provides intent annotations along with natural language explanations from human drivers/annotators. These verbal justifications offer unique insight into the cognitive and contextual cues shaping pedestrian behavior interpretation.

To this end, we developed a feature extraction pipeline that integrates linguistic cues, numerical features, and cross-modal interactions into four semantically meaningful streams:
\begin{itemize}
    \item \textit{\textbf{Psychological attention features (attention features)}}: gaze direction, head orientation, hesitation, and commitment cues extracted from textual descriptions; (e.g., \textit{looking at vehicle}).
    \item \textit{\textbf{Positional features}}: spatial relationships such as curb proximity, crosswalk location, and relative pedestrian-vehicle positioning; (e.g.,\textit{distance to the vehicle}).
    \item \textit{\textbf{Situational features}}: describing traffic context, including vehicle motion, scene density, and nearby agents; (e.g., 
    \textit{nearest vehicle speed}, \textit{local vehicle density}).
    \item \textit{\textbf{Interaction features}}: pairwise products across the other three streams, representing combined behaviors. (e.g., \textit{looking while stepping}, \textit{hesitating while vehicle slows}).
\end{itemize}

This structured representation is inspired by how drivers reason about crossing intention. The temporally aligned are fed to the model for prediction. Although it does not process full-frame sequences, temporal cues are captured implicitly through features aggregated over short time windows, including pedestrian motion, visual scanning behavior, and ego-vehicle dynamics. Because PSI’s dataset is sparse, full temporal encoders (e.g., LSTMs) would learn from noisy sequence targets; therefore, short-window, semantically enriched features provide a more stable representation.

\section{Model Architecture}
\label{sec:model}
Our proposed architecture (Fig.~\ref {fig:architecture}) maps structured pedestrian–vehicle interaction features into a unified representation for intention prediction. The four input streams (Psychological attention, positional, situational, and interaction) are first encoded by independent highway networks and tagged with a learnable stream embedding. A multi-layer Transformer encoder then captures temporal and cross-stream dependencies. Global self-attention pooling yields a single latent vector that feeds a residual MLP classifier producing the crossing probability. Auxiliary KL divergence and Mahalanobis distance heads estimate uncertainty to enhance interpretability. We describe each component in the following subsections.

\subsection{Input Structure}
For each pedestrian instance, we extract three primary input streams: 
$\mathbf{x}^{(a)} \in \mathbb{R}^{d_a}$ (attention features),
$\mathbf{x}^{(p)} \in \mathbb{R}^{d_p}$ (positional features), and
$\mathbf{x}^{(s)} \in \mathbb{R}^{d_s}$ (situational features).

As part of the situational stream, we include semantic text embeddings computed using the paraphrase-mpnet-base-v2 model from SentenceTransformers \cite{reimers2019sentence, song2020mpnet}. In addition, we construct a fourth stream of interaction features, defined as $\mathbf{x}^{(i)} \in \mathbb{R}^{d_i}$, where each element is a pairwise products across attention, positional, and situational features (e.g., $x_k^{(a)} \cdot x_l^{(p)}$ or $x_k^{(s)} \cdot x_l^{(a)}$). All features are independently normalized using RobustScaler and then fitted on the training split to prevent leakage \cite{pedregosa2011scikit}.

\subsection{Highway Encoders}
Each feature stream $\mathbf{x}^{(j)}$ ($j \in \{a,p,s,i\}$ for attention, positional, situational, and interaction features) is processed independently by a dedicated encoder based on a highway architecture \cite{srivastava2015training}. The encoder allows the model to adaptively balance between transforming the input and preserving its original form to improve gradient flow and interpretation.

\noindent Given a feature vector $\mathbf{x}^{(j)}$ from stream j, the encoder computes its output $\mathbf{h}^{(j)}$ as a gated combination of a nonlinear transformation and a passthrough projection:

\begin{equation}
\mathbf{h}^{(j)} \;=\; \mathbf{g}^{(j)} \odot T(\mathbf{x}^{(j)}) \;+\; \bigl(1-\mathbf{g}^{(j)}\bigr) \odot P(\mathbf{x}^{(j)})
\end{equation}
\begin{equation}
\mathbf{g}^{(j)}=\sigma\!\big(W_g^{(j)}\mathbf{x}^{(j)}+\mathbf{b}_g^{(j)}\big)
\end{equation}

\noindent where $W_g^{(j)}$ is the learnable gate matrix,
$T(\cdot)$ is a two-layer feedforward network with GELU activation \cite{hendrycks2016gaussian}, layer normalization \cite{ba2016layer}, and dropout \cite{srivastava2014dropout},
$P(\cdot)$ is a linear projection used for mismatched input–output dimensions, and
$\sigma(\cdot)$ is the sigmoid gating function. This architecture enables the model to selectively transform or preserve features, improving interpretability and gradient flow across deep layers.

\subsection{Stream Embedding and Feature Fusion}
To fuse heterogeneous input modalities, we add learnable stream embeddings that encode the identity of each feature stream. These embeddings help the Transformer distinguish between different behavioral sources during self-attention.

\noindent Let $\mathbf{h}^{(j)} \in \mathbb{R}^{B \times d}$ denote the output of the highway encoder for stream $j$, where $B$ is the batch size and $d$ is the hidden dimension. We associate with each stream a learnable embedding vector $\mathbf{e}^{(j)} \in \mathbb{R}^{d}$, initialized randomly and optimized during training to preserve stream identity.


\noindent The embedded stream representations are computed as:

\begin{equation}
 \overline{\mathbf{h}}^{(j)} = \mathbf{h}^{(j)} + \mathbf{e}^{(j)}.
\end{equation}

\noindent This additive embedding provides stream-specific context to the Transformer, similar in spirit to segment embeddings in BERT \cite{devlin2019bert}.

\noindent The resulting vectors $\overline{\mathbf{h}}^{(a)}$, $\overline{\mathbf{h}}^{(p)}$, $\overline{\mathbf{h}}^{(s)}$, and $\overline{\mathbf{h}}^{(i)}$ are concatenated into $\mathbf{H} = [\overline{\mathbf{h}}^{(a)};\overline{\mathbf{h}}^{(p)};\overline{\mathbf{h}}^{(s)};\overline{\mathbf{h}}^{(i)}] \in \mathbb{R}^{B \times 4 \times d}$.

\noindent We use the fixed order $(a,p,s,i)$ and therefore do not apply a separate positional encoding.

\subsection{Transformer Encoder and Global Attention}
To capture inter-stream dependencies, we employ a multi-layer Transformer encoder \cite{vaswani2017attention} followed by a multi-head self-attention layer that further contextualizes the four stream tokens; we then flatten them into a single vector.

\noindent The embedded sequence $\mathbf{H} \in \mathbb{R}^{B \times 4 \times d} $, consisting of encoded attention, positional, situational, and interaction tokens, is passed through a Transformer encoder with $L$ layers. Each layer is pre-normalized and comprises multi-head self-attention and a GELU-activated feed-forward sublayer with residual connections and dropout:

\begin{equation}
\mathbf{H}'= \text{TransformerEncoder} (\mathbf{H}) \in \mathbb{R}^{B \times 4 \times d}
\end{equation}

\noindent Self-attention within each layer follows \cite{vaswani2017attention}:

\begin{equation}
\text{Attention}(\mathbf{Q}, \mathbf{K}, \mathbf{V}) = \text{softmax}\left( \frac{\mathbf{Q}\mathbf{K}^\top}{\sqrt{d_k}} \right)\mathbf{V}
\end{equation}

\noindent where $Q$,$K$, and $V$ are learned linear projections of the input to the attention block, with $h$ heads and per-head dimension $d_k = \frac{d}{h}$.

\noindent To globally re-weight and mix information across the four tokens, we apply cross-stream multi-head self-attention where $\mathbf{Q} = \mathbf{K} = \mathbf{V} = \mathbf{H}'$ \cite{vaswani2017attention}. This computes attention across the four semantic tokens and produces contextualized representations that are then flattened into a single vector:

\begin{equation}
\mathbf{Z}, \mathbf{A} = \text{MultiHeadAttention}(\mathbf{H}', \mathbf{H}', \mathbf{H}')
\end{equation}

\noindent where $\mathbf{Z} \in \mathbb{R}^{B \times 4 \times d}$ contains attention-weighted representations of the four streams, and $\mathbf{A} \in \mathbb{R}^{B \times 4 \times 4}$ denotes the head-averaged cross-stream attention weights for interpretation.

\noindent Finally, we flatten the four contextualized tokens to form the fused representation:

\begin{equation}
\mathbf{z} = \text{Flatten}(\mathbf{Z}) \in \mathbb{R}^{B \times 4d}
\end{equation}
which is fed to the downstream classifier and anomaly components.

\subsection{Classification Head}
Following contextual fusion, the flattened output $\mathbf{z} \in \mathbb{R}^{B \times 4d}$ is passed through a multi-stage classification head with residual connections designed for binary prediction.

\paragraph{Pre-classification Transformation.}
We first project $\mathbf{z}$ into a lower-dimensional space via a linear layer followed by LayerNorm, GELU, and Dropout:

{\small
\begin{equation}
\mathbf{z}_0 \;=\; \mathrm{Dropout}\!\big(\mathrm{GELU}(\mathrm{LayerNorm}(W_0 \mathbf{z} + \mathbf{b}_0))\big), 
 \mathbf{z}_0 \in \mathbb{R}^{B \times 3d}.
\end{equation}
}

\paragraph{Residual MLP Stack.}
The transformed vector is then passed through three residual blocks that progressively reduce the dimensionality. Each block performs a linear transformation followed by residual addition:

{\small
\begin{equation}
\mathbf{z_{k+1}} = \text{Dropout}(\text{GELU}(\text{LayerNorm}(W_kz_k+b_k)))+P_k\mathbf{z_k}
\end{equation}
}

If input and output dimensions differ, a linear projection aligns the residual connection.

\paragraph{Final Prediction.}
The output of the final residual block is projected through a two-layer prediction head:

{\small
\begin{equation}
\mathbf{z}_{\mathrm{cls}} \;=\; \mathrm{Dropout}\!\big(\mathrm{GELU}(W_1 \mathbf{z}_3 + \mathbf{b}_1)\big),
\end{equation}
\begin{equation}
\mathrm{logit} \;=\; W_2 \mathbf{z}_{\mathrm{cls}} + \mathbf{b}_2 .
\end{equation}
}

\noindent The crossing probability is obtained via a sigmoid activation:

{\small
\begin{equation}
\widehat{y} \;=\; \sigma(\mathrm{logit}) \;=\; \frac{1}{1 + e^{-\mathrm{logit}}},
\end{equation}
}
yielding the crossing probability $\widehat y \in [0,1]$. We train with label smoothing, and the sigmoid is applied only during evaluation; see "Loss Function and Calibration" section for details.

\subsection{Anomaly Detection}
To identify out-of-distribution behaviors and quantify predictive uncertainty, we use two complementary signals: a variational anomaly embedding trained with a KL prior, and a Mahalanobis distance measured in that embedding space.

\paragraph{Variational Anomaly Embedding.}
After cross–stream fusion, a variational bottleneck predicts the mean $\boldsymbol{\mu}\in\mathbb{R}^{d_z}$ and log–variance $\boldsymbol{\eta}=\log\boldsymbol{\sigma}^2\in\mathbb{R}^{d_z}$ from the fused vector $\mathbf{z}$:

\begin{equation}
    \boldsymbol{\mu}=W_\mu \mathbf{z}+ \mathbf{b}_\mu,
\end{equation}
\begin{equation}
\boldsymbol{\eta} = W_{\log\sigma^2}\mathbf{z} + \mathbf{b}_{\log\sigma^2}
\end{equation}

\noindent We then sample the latent vector using the reparameterization trick during training \cite{kingma2013auto,rezende2014stochastic}:

\begin{equation}
\mathbf{z}_{\text{anom}} \;=\; \boldsymbol{\mu} \;+\; \exp\!\big(\tfrac{1}{2}\boldsymbol{\eta}\big)\odot \boldsymbol{\epsilon},
\qquad \boldsymbol{\epsilon}\sim\mathcal{N}(\mathbf{0},\mathbf{I}_{d_z}).
\end{equation}

\noindent The Kullback–Leibler divergence to the unit Gaussian prior provides a per-sample regularizer \cite{kingma2013auto}:

\begin{equation}
\mathcal{L}_{\text{KL}} \;=\; -\tfrac{1}{2}\sum_{i=1}^{d_z}\Big(1 + \eta_i - \mu_i^2 - e^{\eta_i}\Big).
\end{equation}

\noindent At evaluation, we use the deterministic mean embedding ($\mathbf{z}_{\text{anom}}=\boldsymbol{\mu}$). We map the KL value through a sigmoid, $\text{score}_{\text{KL}}=\sigma(\mathcal{L}_{\text{KL}})$, so larger deviations yield higher anomaly scores. The KL term’s weight $\beta$ is small and mildly increased over training.

\paragraph{Mahalanobis Distance Detector.}
We detect embedding-level outliers via the squared Mahalanobis distance \cite{mahalanobis1936,de2000mahalanobis}. 
Let $\{\boldsymbol{\mu}_n\}$ be the mean embeddings on the training set. We estimate the training centroid and a shrinkage covariance using the Ledoit–Wolf estimator \cite{ledoit2004well}, and add a small ridge term for numerical stability.

\begin{equation}
\hat{\boldsymbol{\mu}}=\tfrac{1}{N}\sum_n \boldsymbol{\mu}_n,\qquad 
\end{equation}
\begin{equation}
\hat{\boldsymbol{\Sigma}}=\text{LW}\!\big(\{\boldsymbol{\mu}_n-\hat{\boldsymbol{\mu}}\}\big)+\lambda \mathbf{I}.
\end{equation}

\noindent For a test sample with mean embedding $\boldsymbol{\mu}$, the distance is
\begin{equation}
D_M^2(\boldsymbol{\mu})=(\boldsymbol{\mu}-\hat{\boldsymbol{\mu}})^\top \hat{\boldsymbol{\Sigma}}^{-1}(\boldsymbol{\mu}-\hat{\boldsymbol{\mu}}).
\end{equation}

\noindent This complements the KL-based signal: KL measures deviation from the prior, while Mahalanobis measures deviation from the training distribution. Both are computed per sample at inference.

\subsection{Loss Function and Calibration}
The model is trained with a composite objective that balances classification accuracy, variational regularization, feature diversity, and a mild temperature prior. The total loss is:

\begin{equation}
\mathcal{L}_{\text{total}}
= \mathcal{L}_{\text{CE}}
+ \lambda_{\text{KL}}\mathcal{L}_{\text{KL}}
+ \lambda_{\text{div}}\mathcal{L}_{\text{div}}
+ \lambda_{\tau}\lvert \tau - 1\rvert .
\end{equation}

\noindent\textbf{Cross-entropy with MixUp and label smoothing.}
For each mini-batch we apply MixUp to inputs and labels,
\begin{equation}
    y^{mix} = \lambda y + (1- \lambda)y_\pi
\end{equation}
with $\lambda \sim Beta(0.2,0.2)$, then smooth target as:
\begin{equation}
    t = (1- \epsilon) y^{mix}+\frac{\epsilon}{2}
\end{equation}
and optimize binary cross-entropy with logits:

\begin{equation}
\mathcal{L}_{\text{CE}}
= -\, t \,\log \sigma(\ell)\;-\; (1-t)\,\log\!\big(1-\sigma(\ell)\big),
\end{equation}
where $\ell$ are the pre-sigmoid logits and $\sigma$ is the sigmoid.

\noindent\textbf{Variational regularization (KL).}
$\mathcal{L}_{\text{KL}}$ is the KL divergence between the variational posterior and the unit Gaussian prior (see Anomaly Detection). We apply a small, mildly annealed coefficient $\lambda_{\text{KL}}$ over training.

\noindent\textbf{Feature diversity penalty.}
To discourage representational collapse, we penalize low batchwise variance in the intermediate features $F\in\mathbb{R}^{B\times m}$:

\begin{equation}
\mathcal{L}_{\text{div}} \;=\; -\,\frac{1}{m}\sum_{j=1}^{m}\operatorname{Std}_b\!\left(F_{b,j}\right).
\end{equation}

Its weight $\lambda_{\text{div}}$ is increased slightly during training.

\noindent\textbf{Temperature prior.}
We encourage the learnable temperature to remain near unity via $\mathcal{L}_{\tau}=\lvert \tau-1\rvert$ with a small fixed coefficient $\lambda_{\tau}$.

\noindent\textbf{Temperature calibration.}
At evaluation time, we calibrate probabilities by applying the learned temperature to the logits:

\begin{equation}
\hat{y}_{\text{cal}} \;=\; \sigma\!\left(\frac{\ell}{\tau}\right),
\end{equation}

where $\ell$ denotes the pre-sigmoid logit and $\sigma$ is the sigmoid \cite{guo2017calibration}. The temperature is not used inside the training loss (aside from the prior above) and is only applied at inference.
The model is lightweight ($\approx1–2M$ parameters) and uses no video backbone at inference. Because PSI provides structured features, end-to-end latency depends on upstream perception modules, making hardware-specific runtime outside our scope.

\section{Results and Analysis}

\subsection{Experimental Setup}
We evaluate on \textbf{PSI~1.0} and \textbf{PSI~2.0}, using the official train/validation/test splits. PSI-1.0 contains 110 videos, and PSI-2.0 extends this to 204 videos. Both versions share the same annotation schema and feature definitions; PSI-2.0 simply provides more videos and annotations for broader evaluation \cite{psi_github}. Preprocessing follows the Input Structure procedure: per-stream scaling is fit on training data, semantic embeddings and interaction features are constructed as described, and streams are concatenated in the order $(a, p, s, i)$. Training runs for 220 epochs with AdamW, including a 15-epoch linear warm-up followed by cosine decay. Regularization includes dropout, label smoothing, MixUp (feature and label blending), weighted sampling for class imbalance, and gradient clipping. We report Accuracy, F1-score, AUC-ROC, and Average Precision. Decision thresholds were chosen on the validation set to maximize an F1-dominant metric, and the selected value was fixed for test evaluation. Per-class Precision, Recall, and F1 are also included. At evaluation, we apply post-hoc temperature scaling on validation logits for calibrated probabilities. For uncertainty, we report two primary anomaly scores: a KL-based score from the variational head and the Mahalanobis distance of the latent embedding to the training distribution.

\subsection{Overall Classification Performance}

We evaluate our model on \textbf{PSI~1.0} using the official train/validation/test split.  
Table~\ref{tab:psi1_comparison} compares our approach with a range of intent-prediction models\cite{Uziel_2025_WACV}, including two recent strong baselines, ClipCross~\cite{Uziel_2025_WACV} and TrEP~\cite{zhang2023trep}.  
While these methods use vision–language embeddings, our approach uses compact, interpretable behavioural features; the comparison is thus modality-agnostic and intended only to contextualize benchmark performance.  
Even so, our model achieves higher test Accuracy, F1, AUC-ROC, and MCC. Results are reported as mean ± standard deviation over five seeds, with thresholds selected on the validation set and applied unchanged to the test set.  
Given the lightweight nature of our feature stream, these results suggest that structured behavioural cues and vision–language embeddings are complementary, and may be effectively combined in future work.

\begin{table}[h]
\centering
\caption{Performance comparison on the PSI 1.0 test set \\
(baseline numbers are reported from \cite{Uziel_2025_WACV}).}
\label{tab:psi1_comparison}
\begin{tabular}{lcccc}
\toprule
\textbf{Method} & \textbf{Accuracy} & \textbf{F1} & \textbf{AUC-ROC} & \textbf{MCC}\\
\midrule
SF-GRU         & 0.788 & 0.719 & 0.752 & 0.452 \\
SingleRNN      & 0.782 & 0.714 & 0.734 & 0.440 \\
MultiRNN       & 0.658 & 0.611 & 0.666 & 0.229 \\
PCPA           & 0.682 & 0.584 & 0.611 & 0.176 \\
ARN            & 0.688 & 0.618 & 0.671 & 0.237 \\
PSI            & 0.759 & 0.681 &   -   & 0.374 \\
TrEP           & 0.830 & 0.771 &   -   & 0.561 \\
ClipCross      & 0.830 & 0.795 & 0.855 & 0.591 \\
\textbf{Ours(PSI1)}  & \textbf{0.88 ± 0.01} & \textbf{0.90 ± 0.01} & \textbf{0.94 ± 0.01} & \textbf{0.78 ± 0.02}\\
\bottomrule
\end{tabular}
\end{table}

We conducted ablations on PSI~1.0 by keeping the architecture and training setup fixed and removing each behavioral stream in turn (Table~\ref{tab:psi1_ablationstudy}).
Removing the attention or positional streams causes modest F1 declines ($0.90\rightarrow0.88$ and $0.90\rightarrow0.86$), confirming that gaze and spatial layout offer complementary cues. In contrast, removing the situational stream sharply reduces performance ($0.90\rightarrow0.75$, MCC $0.77\rightarrow0.45$), which underscores the importance of contextual information. Disabling the interaction stream slightly improves results, which suggests redundancy with the Transformer’s learned fusion.
Overall, the model is not dominated by any single stream, and situational cues are most influential, and others enhance robustness. Since ablations target component analysis, we report them only on PSI~1.0. PSI~2.0 focuses on cross-version generalization and is reported without ablation. 

\begin{table}[h]
\centering
\caption{Ablation study of behavioral streams and training components on PSI 1.0.}
\label{tab:psi1_ablationstudy}
\begin{tabular}{lcccc}
\toprule
\textbf{Ablation Setting} & \textbf{Accuracy} & \textbf{F1} & \textbf{AUC-ROC} & \textbf{MCC} \\
\midrule
Full Model (Ours) & 0.88 & 0.90 & 0.94 & 0.77 \\
w/o Attention stream & 0.86 & 0.88 & 0.93 & 0.73 \\
w/o Positional stream & 0.85 & 0.86 & 0.94 & 0.73 \\
w/o Situational stream & 0.73 & 0.75 & 0.79 & 0.45 \\
w/o Interaction stream & 0.89 & 0.90 & 0.94 & 0.78 \\
\bottomrule
\end{tabular}
\end{table}

To the best of our knowledge (as of August~2025), no published work has yet reported results on \textbf{PSI~2.0}, which is a more diverse extension of the original dataset. Using the same training protocol, our model achieves \textbf{F1 0.781 ± 0.013}, \textbf{AUC 0.793 ± 0.009}, and \textbf{MCC 0.440 ± 0.021} (Table~\ref{tab:psi2_results}), establishing an initial baseline for future PSI~2.0 studies.

\begin{table}[h]
\centering
\caption{PSI 2.0 test performance (mean ± std over five seeds; threshold chosen on validation and kept fixed for test).}
\label{tab:psi2_results}
\begin{tabular}{lcccc}
\toprule
\textbf{Method} & \textbf{Accuracy} & \textbf{F1} & \textbf{AUC-ROC} & \textbf{MCC} \\
\midrule
\textbf{Ours} & \textbf{0.73 ± 0.01} & \textbf{0.78 ± 0.01} & \textbf{0.79 ± 0.01} & \textbf{0.44 ± 0.02} \\
\bottomrule
\end{tabular}
\end{table}

\subsection{Per-Class Analysis}

Table~\ref{tab:per_class_perf} reports per-class precision, recall, and F1 on the test sets.  
On \textbf{PSI~1.0}, the model achieves balanced performance, with F1 scores of 0.868 for the non-crossing class and 0.893 for the crossing class, which outperforms prior work on both.  
On the more challenging \textbf{PSI~2.0} benchmark, the model maintains high recall (0.812) and F1 (0.781) on the safety-critical crossing class, while precision remains acceptable at 0.757.  
As expected, performance on the majority non-crossing class is lower (F1 = 0.648); however, the model maintains high precision (0.698) and favors a conservative strategy of flagging potential crossings, which is a common design tradeoff in pedestrian intent prediction.

\begin{table}[h]
\centering
\caption{Per-class precision, recall, and F1 (mean over five seeds).}
\label{tab:per_class_perf}
\begin{tabular}{llccc}
\toprule
\textbf{Dataset} & \textbf{Class} & \textbf{Precision} & \textbf{Recall} & \textbf{F1} \\
\midrule
\multirow{2}{*}{PSI 1.0}
 & Non-Crossing (0) & 0.805 & 0.943 & 0.868 \\
 & Crossing (1)     & 0.955 & 0.840 & 0.893 \\
\midrule
\multirow{2}{*}{PSI 2.0}
 & Non-Crossing (0) & 0.698 & 0.614 & 0.648 \\
 & Crossing (1)     & 0.757 & 0.812 & 0.781 \\
\bottomrule
\end{tabular}
\end{table}

\subsection{Anomaly Detection and Uncertainty Analysis}

To evaluate the reliability and robustness of our model, we analyze uncertainty and anomaly detection using following complementary approaches:

\subsubsection{Uncertainty Calibration}
Table~\ref{tab:calib} summarizes calibration quality before and after applying the learnable temperature parameter $\tau$.
We assess calibration using Brier score, Expected Calibration Error (ECE), and negative log-likelihood (NLL), both on raw logits and after post-hoc temperature scaling~\cite{guo2017calibration} with $\tau$ fitted on the validation set. On \textbf{PSI 1.0}, temperature scaling changes little and can slightly worsen NLL/ECE, indicating the base model is already near-calibrated. On \textbf{PSI 2.0}, it yields a clear improvement in all three metrics.

\begin{table}[h]
\centering
\caption{Calibration on test sets (mean $\pm$ std over 5 seeds)}
\label{tab:calib}
\vspace{0.3em}
\begin{tabular}{lccc}
\toprule
\textbf{Dataset / Setting} & \textbf{Brier} $\downarrow$ & \textbf{ECE} $\downarrow$ & \textbf{NLL} $\downarrow$ \\
\midrule
PSI 1.0 (raw)            & $0.093 \pm 0.004$ & $0.360 \pm 0.033$ & $0.320 \pm 0.020$ \\
PSI 1.0 ($\tau$-scaled)  & $0.093 \pm 0.004$ & $0.367 \pm 0.031$ & $0.333 \pm 0.028$ \\
\midrule
PSI 2.0 (raw)            & $0.214 \pm 0.012$ & $0.311 \pm 0.026$ & $0.753 \pm 0.079$ \\
PSI 2.0 ($\tau$-scaled)  & $0.188 \pm 0.005$ & $0.189 \pm 0.031$ & $0.562 \pm 0.011$ \\
\bottomrule
\end{tabular}
\end{table}

Overall, temperature scaling brings \emph{substantial} gains on PSI~2.0 (ECE $0.311 \!\to\! 0.189$, NLL $0.753 \!\to\! 0.562$, Brier $0.214 \!\to\! 0.188$), while its effect on PSI~1.0 is neutral to slightly negative. For completeness, when ECE is computed using predicted-class confidence (rather than class-1 probability), it is much smaller on both datasets (PSI~1.0: $0.050 \pm 0.020$, PSI~2.0: $0.037 \pm 0.018$), indicating that chosen-class confidences are well aligned with empirical accuracy; the larger class-probability ECE partly reflects class imbalance.



\begin{table}[h]
\centering
\caption{Error-detection performance (mean $\pm$ std over 5 seeds). Higher is better.}
\label{tab:err_det}
\begin{tabular}{lcccc}
\toprule
\multirow{2}{*}{\textbf{Dataset}} & \multicolumn{2}{c}{\textbf{AUROC}} & \multicolumn{2}{c}{\textbf{AUPRC}} \\
\cmidrule(lr){2-3}\cmidrule(lr){4-5}
& KL $\uparrow$ & Mahalanobis $\uparrow$ & KL $\uparrow$ & Mahalanobis $\uparrow$ \\
\midrule
PSI~1.0 & $0.59 \pm 0.08$ & $\mathbf{0.77 \pm 0.03}$ & $0.17 \pm 0.04$ & $\mathbf{0.33 \pm 0.09}$ \\
PSI~2.0 & $0.53 \pm 0.06$ & $\mathbf{0.66 \pm 0.02}$ & $0.33 \pm 0.04$ & $\mathbf{0.43 \pm 0.03}$ \\
\bottomrule
\end{tabular}
\end{table}

\begin{table}[h]
\centering
\caption{Accuracy after abstaining on the riskiest samples (Mahalanobis \emph{class-conditional}). Mean $\pm$ std over 5 seeds.}
\label{tab:risk_cov}
\begin{tabular}{lccc}
\toprule
\textbf{Dataset} & \textbf{100\% (all)} & \textbf{90\% kept} & \textbf{80\% kept} \\
\midrule
PSI~1.0 & $0.888 \pm 0.007$ & $\mathbf{0.914 \pm 0.007}$ & $\mathbf{0.935 \pm 0.006}$ \\
PSI~2.0 & $0.708 \pm 0.027$ & $\mathbf{0.731 \pm 0.031}$ & $\mathbf{0.752 \pm 0.026}$ \\
\bottomrule
\end{tabular}
\end{table}

\subsubsection{Error Detection with KL and Mahalanobis}

To assess whether the uncertainty heads can flag unreliable predictions, we label each test example as \emph{correct} or \emph{incorrect} and measure how well the \textbf{KL‐based} and \textbf{Mahalanobis} scores separate the two groups.
Table~\ref{tab:err_det} reports AUROC and AUPRC, averaged over the five random seeds used throughout our experiments.

\begin{figure}[h]
\centering
\includegraphics[width=0.83\linewidth]{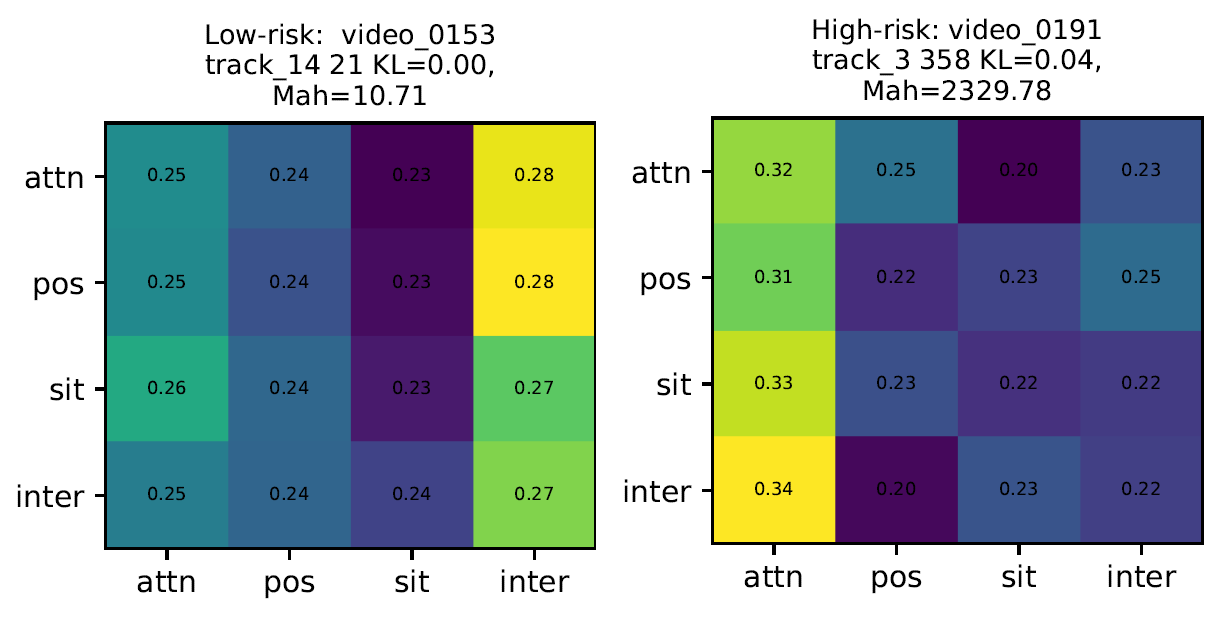}
\caption{Cross-stream attention heat-maps and uncertainty scores for two PSI~2.0 samples. Each row shows average multi-head attention from one stream token (attn/pos/sit/inter) to all streams.
Left: low model risk (KL\,=\,0.00, Mah\,=\,10.71). 
Right: high model risk (KL\,=\,0.04, Mah\,=\,2329.78). 
Cell values are mean attention weights across the four behavioral streams.}
\label{fig:qualitative}
\end{figure}

\begin{figure*}[h]
    \centering
    
    \begin{subfigure}{0.24\textwidth}
        \centering
        \includegraphics[width=0.81\linewidth]{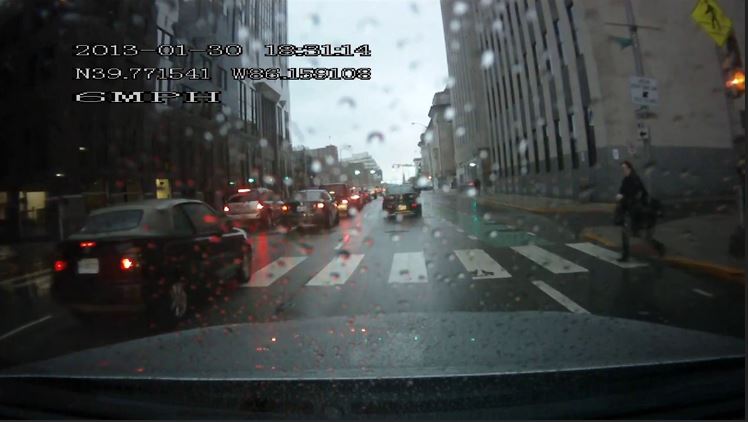}
        \includegraphics[width=0.96\linewidth]{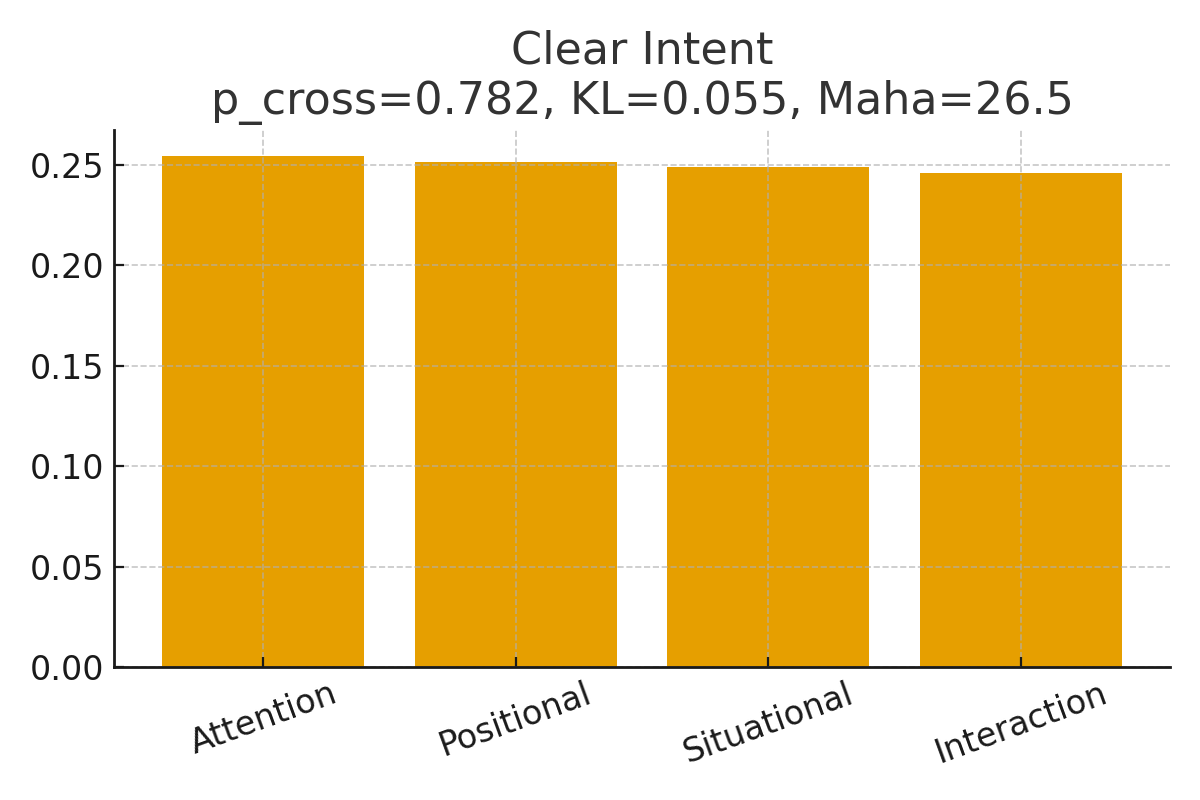}
        \caption{Clear Intent}
    \end{subfigure}
    \hfill
    \begin{subfigure}{0.24\textwidth}
        \centering
        \includegraphics[width=0.81\linewidth]{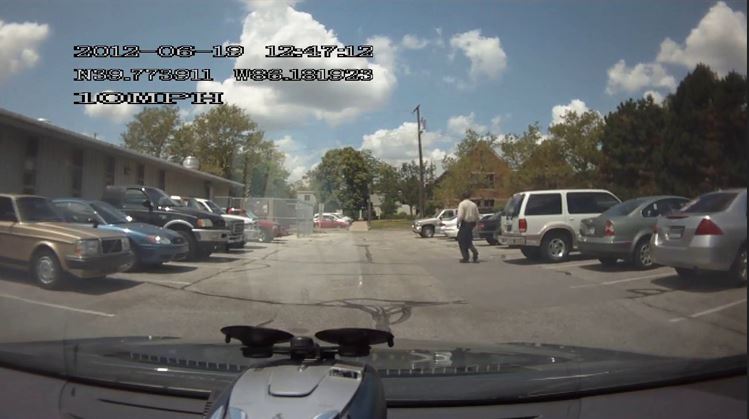}
        \includegraphics[width=0.96\linewidth]{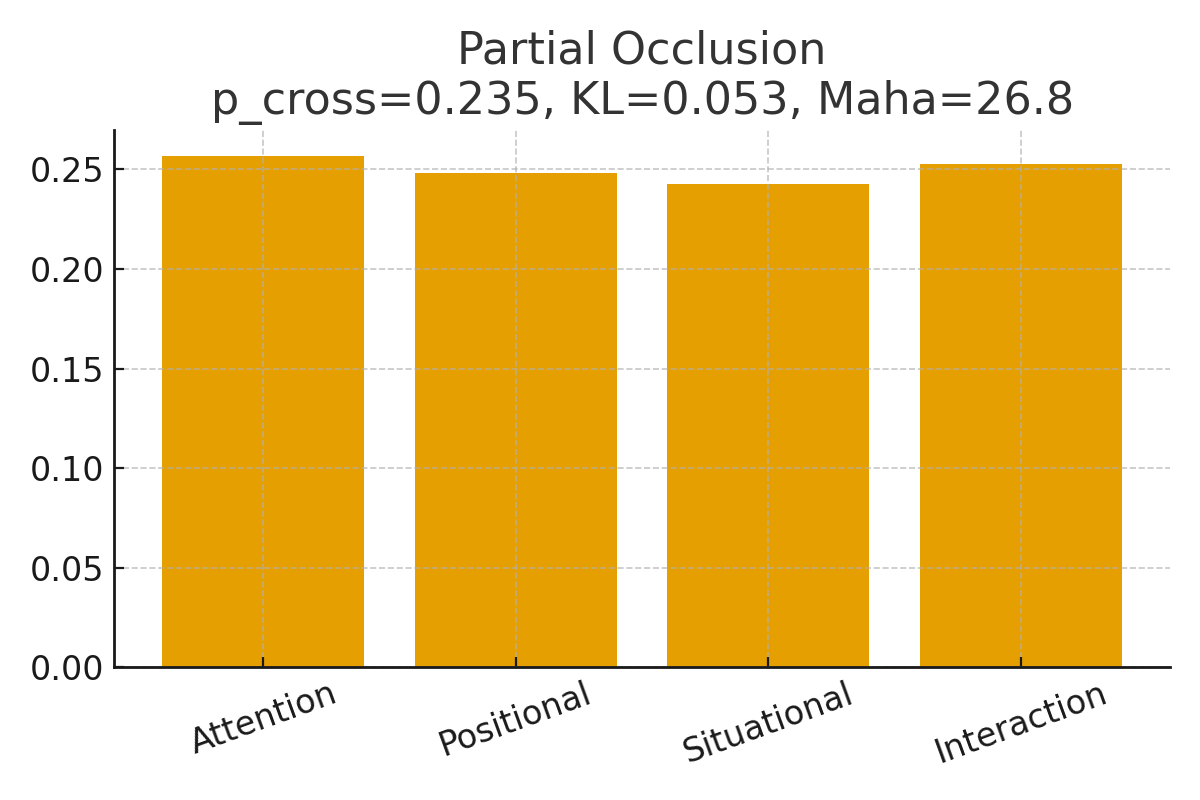}
        \caption{Partial Occlusion}
    \end{subfigure}
    \hfill
    \begin{subfigure}{0.24\textwidth}
        \centering
        \includegraphics[width=0.81\linewidth]{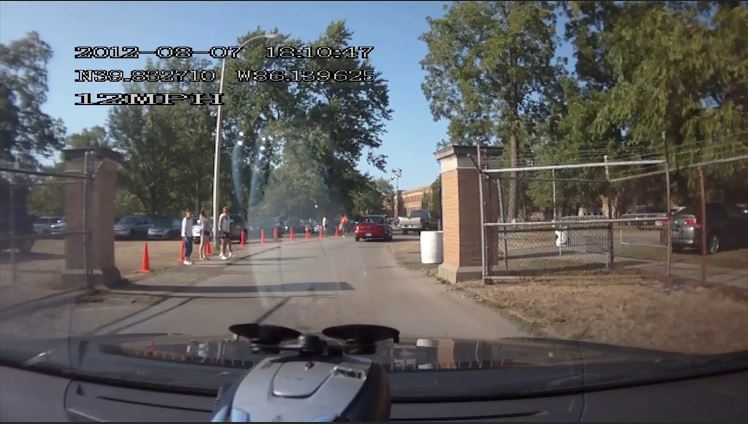}
        \includegraphics[width=0.96\linewidth]{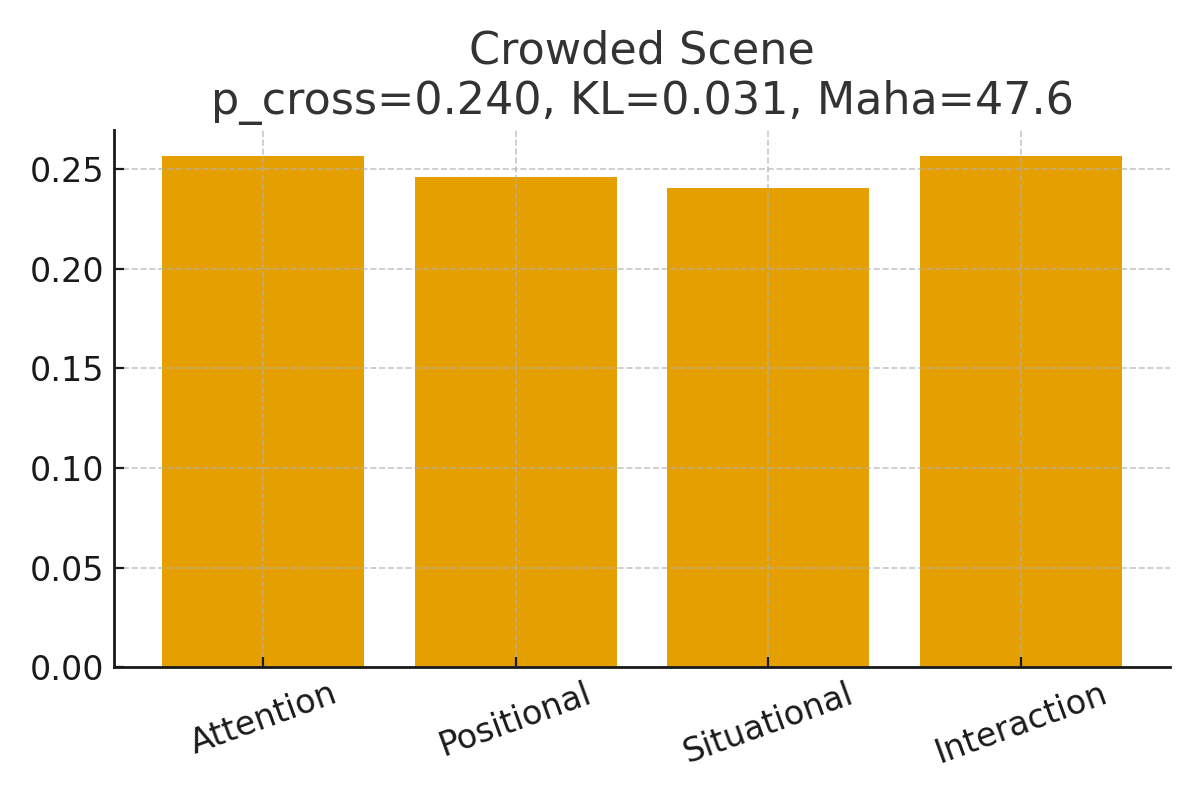}
        \caption{Crowded Scene}
    \end{subfigure}
    \hfill
    \begin{subfigure}{0.24\textwidth}
        \centering
        \includegraphics[width=0.81\linewidth]{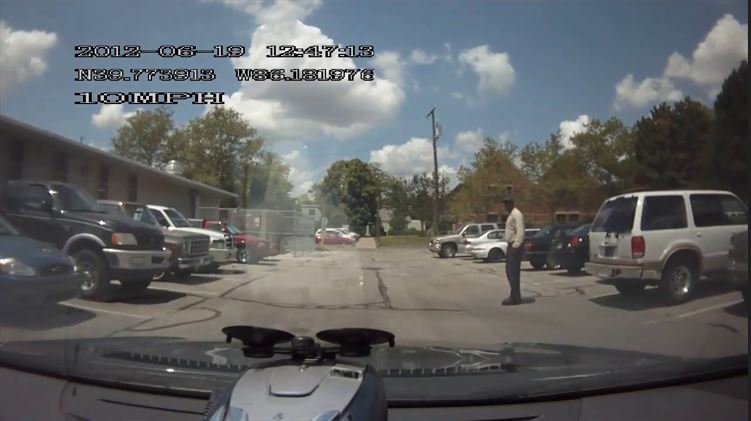}
        \includegraphics[width=0.96\linewidth]{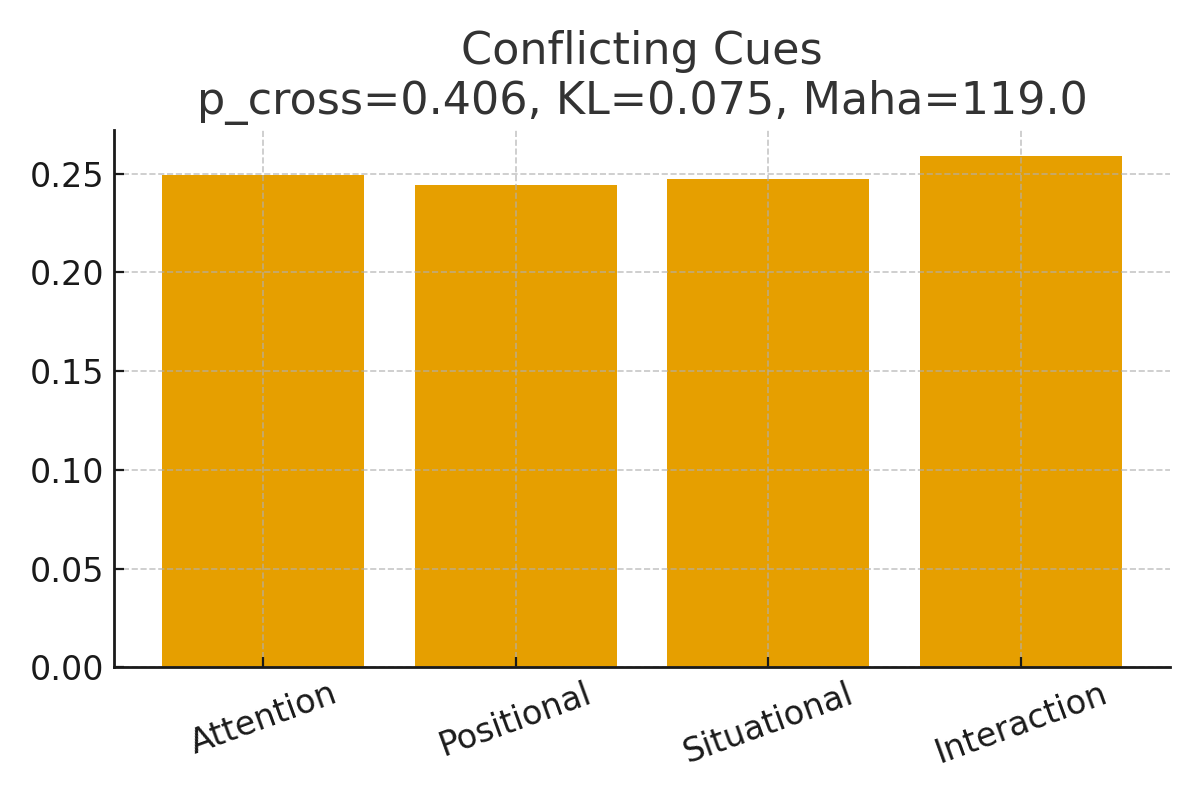}
        \caption{Conflicting Cues}
    \end{subfigure}

    \caption{Four composite qualitative examples showing input frames with a visualization of the behavioural-stream attention weights and uncertainty estimates (KL and Mahalanobis).}
    \label{fig:qualitative_composite}
\end{figure*}

On both \textbf{PSI~1.0} and \textbf{PSI~2.0}, the \emph{class-conditional Mahalanobis} distance computed in the classifier feature space is the strongest single error indicator, clearly outperforming the \emph{KL-based} score in both AUROC and AUPRC.%
\footnote{We also evaluated margin/entropy confidence and a simple rank-ensemble; both are competitive but still trail class-conditional Mahalanobi.}

\noindent
\textbf{Selective prediction.} 
Ranking test samples by \emph{class-conditional Mahalanobis} and retaining only the lowest-risk cases improves accuracy at fixed coverage on both datasets (Table~\ref{tab:risk_cov}). On \textbf{PSI~1.0}, accuracy rises from $0.888 \pm 0.007$ (all samples) to $0.914 \pm 0.007$ at $90\%$ kept and $0.935 \pm 0.006$ at $80\%$ kept. On \textbf{PSI~2.0}, accuracy increases from $0.708 \pm 0.027$ to $0.731 \pm 0.031$ ($90\%$ kept) and $0.752 \pm 0.026$ ($80\%$ kept).%
\footnote{For reference, using the \emph{standard} (class-agnostic) Mahalanobis score yields $0.901 \pm 0.010$ ($90\%$) and $0.922 \pm 0.009$ ($80\%$) on PSI~1.0; $0.724 \pm 0.031$ ($90\%$) and $0.745 \pm 0.038$ ($80\%$) on PSI~2.0.}

\noindent
\textbf{\emph{Interpreting AUPRC.}} The positive class for error detection is ``incorrect'' predictions, whose prevalence equals the error rate. Random AUPRC is therefore the error rate (\(\approx\)~$0.11$ on PSI~1.0; \(\approx\)~$0.29$ on PSI~2.0). Our best AUPRCs are $\sim\!3\times$ random on PSI~1.0 and $\sim\!1.5\times$ random on PSI~2.0, indicating practically useful error ranking despite modest absolute values.

\noindent
\textbf{Selective prediction (takeaway).} Even a simple “abstain if risk $>$ threshold” policy using class-conditional Mahalanobis yields clear gains: on \textbf{PSI~1.0}, accuracy rises from $0.888$ (all) to $0.914$ at $90\%$ kept (+2.6 pp) and $0.935$ at $80\%$ kept (+4.7 pp); on \textbf{PSI~2.0}, from $0.708$ to $0.731$ (+2.3 pp) and $0.752$ (+4.4 pp), respectively—providing actionable safety margins for risk-aware decision-making.

\subsection{Qualitative Insights}
Fig.~\ref{fig:qualitative} shows cross-stream attention and uncertainty for one low- and one high-risk PSI~2.0 sample, where “risk’’ denotes \emph{model error likelihood}. Each attention row is softmax-normalized (summing to 1, typically $\approx0.25$ per token).

\begin{itemize}
    \item \textbf{Low-risk example (PSI~2.0):} \emph{video\_0153, track\_14\_21}. 
      A woman crosses on a marked crosswalk while the ego vehicle is stopped (rainy conditions).
      Scene cues are consistent (crosswalk, in-front geometry, ego speed $\approx 0$), yielding $KL = 0.00$ and $Mahalanobis = 10.7$.
      The $4{\times}4$ attention matrix is nearly uniform ($0.23{\sim}0.28$) with slight emphasis on the \emph{interaction} column, indicating routine, geometry-driven reasoning.

    \item \textbf{High-risk example (PSI~2.0):} \emph{video\_0191, track\_3\_358}. 
      A crowded scene ($\approx16$ agents; $12$ vehicles, $4$ pedestrians, and one vehicle is on the crossing path) without language annotations, so the model relies solely on numeric cues.
      $Mahalanobis = 2329.8$ and $KL = 0.04$ highlight complementary heads: the embedding is far from training data, though the posterior is sharp.
      Attention concentrates on the \emph{attention} stream ($max \approx 0.32–0.34$) with lower weights on others (0.20–0.25), marking this sample as high error risk under the selective-prediction policy.

\end{itemize}

These examples show that (i) with four stream tokens, attention is generally diffuse ($\approx0.25$ per row) but small, systematic deviations are informative. Familiar scenes lean toward \emph{positional/interaction} tokens, while atypical cases emphasize \emph{attention}—and (ii) Mahalanobis and KL provide complementary uncertainty cues.
Together, they enable a risk-aware policy that retains 80–90\% of samples while preserving PSI~1.0 accuracy and improving PSI~2.0 results.


\subsection{Qualitative Analysis of Attention and Uncertainty}
To illustrate model reasoning, we analyze four representative PSI~2.0 scenarios—(a) clear intent, (b) partial occlusion, (c) crowded / multi-agent, and (d) conflicting cues, which are selected using joint criteria over predicted probability, KL divergence, Mahalanobis distance, and scene context (Fig.~\ref{fig:qualitative_composite}).

\subsubsection{Clear Intent}
Fig.~\ref{fig:qualitative_composite}(a) shows a clear crossing case (video\_0190, frame 165): a pedestrian walks laterally across a marked crosswalk while the ego vehicle is stopped. The model outputs high confidence ($p=0.78$, $KL = 0.0547$, $Mah = 26.5$) with balanced attention across streams ($\approx 0.25$ each), reflecting unambiguous and consistent cues.

\subsubsection{Partial Occlusion}
Fig.~\ref{fig:qualitative_composite}(b) shows a partially occluded pedestrian (video\_$0147$, frame $139$). With the lower body hidden by parked cars, the model lowers its prediction ($p=0.23$) but maintains low uncertainty ($KL = 0.053$, $Mah = 26.8$). Attention slightly shifts toward interaction and situational cues, compensating for missing visual information.

\subsubsection{Crowded / Multi-Agent Scene}
In Fig.~\ref{fig:qualitative_composite}(c) (video\_0149, frame 33), multiple pedestrians and vehicles create a dense, structured scene. The model predicts low crossing probability ($p=0.24$) and elevated Mahalanobis ($47.6$), increasing weight on situational and interaction streams ($\approx 0.26$) to account for crowd and lane context.

\subsubsection{Conflicting / Ambiguous Cues}
Fig.~\ref{fig:qualitative_composite}(d) (video\_0147, frame\_167) shows a pedestrian paused near parked cars, producing conflicting signals. The model gives $p=0.40$ with high uncertainty ($KL=0.075$, $Mah=119$), attention tilts toward interaction cues ($\approx 0.26$), indicating reliance on relational context but insufficient evidence for a confident decision.

\section{Conclusion}
We presented a socially informed, multi-stream model for pedestrian crossing intent prediction that fuses attention, positional, situational, and interaction cues in a compact Transformer.  On PSI~1.0, it outperforms recent baselines in F1, AUC-ROC, and MCC, and provides a strong baseline for PSI~2.0. Two uncertainty heads (KL divergence and Mahalanobis distance) provide calibrated probabilities, actionable risk estimates, and interpretable attention maps. With its lightweight and modality-agnostic design, the model can complement vision-language models.

\section*{Acknowledgment}
This work was supported in part by the Transportation Network Growth Opportunity (TNGO) initiative funded by the Tennessee Department of Economic and Community Development, in collaboration with the University of Tennessee at Chattanooga and industry partners.

\bibliographystyle{IEEEtran}
\bibliography{ref}

\end{document}